\pdfoutput=1

\documentclass[11pt]{article}

\usepackage[final]{acl}

\usepackage{times}
\usepackage{latexsym}

\usepackage[T1]{fontenc}

\usepackage[utf8]{inputenc}

\usepackage{microtype}

\usepackage{inconsolata}

\usepackage{graphicx}

\usepackage{amsmath}
\usepackage{amssymb}
 
\usepackage{times}
\usepackage{latexsym}
\usepackage{array} 
\usepackage{multirow}

\makeatletter
\let\cas@beginabstract\abstract      
\let\cas@endabstract  \endabstract   
\makeatother

\usepackage{arabtex}
\usepackage{utf8}
\setcode{utf8}

\makeatletter
\let\abstract   \cas@beginabstract   
\let\endabstract\cas@endabstract     
\makeatother

\usepackage{hyperref}  
\usepackage{graphicx}   
\usepackage{amssymb}    
\usepackage{pifont}     
\usepackage{xcolor} 
\newcommand{\xmarkred}{\textcolor{red}{\ding{55}}}
\usepackage[T1]{fontenc}

\usepackage[utf8]{inputenc}

\usepackage{microtype}

\usepackage{inconsolata}

\title{GATE: General Arabic Text Embedding for Enhanced Semantic Textual Similarity with Matryoshka Representation Learning and Hybrid Loss Training}

\author{
\textbf{Omer Nacar\textsuperscript{1*}} \quad
\textbf{Anis Koubaa\textsuperscript{2}} \quad
\textbf{Serry Sibaee\textsuperscript{1}} \\
\textbf{Yasser Al-Habashi\textsuperscript{1}} \quad
\textbf{Adel Ammar\textsuperscript{1}} \quad
\textbf{Wadii Boulila\textsuperscript{1}} \\
\textsuperscript{1}Prince Sultan University, Riyadh, Saudi Arabia \\
\textsuperscript{2}Alfaisal University, Riyadh, Saudi Arabia \\
\texttt{\{onajar, ssibaee, yalhabashi, aammar, wboulila\}@psu.edu.sa}, \texttt{akoubaa@alfaisal.edu.sa} \\
\textsuperscript{*}Corresponding author: \texttt{onajar@psu.edu.sa}
}

\begin{document}
\maketitle

\begin{abstract}
Semantic textual similarity (STS) is a critical task in natural language processing (NLP), enabling applications in retrieval, clustering, and understanding semantic relationships between texts. However, research in this area for the Arabic language remains limited due to the lack of high-quality datasets and pre-trained models. This scarcity of resources has restricted the accurate evaluation and advance of semantic similarity in Arabic text. This paper introduces General Arabic Text Embedding (GATE) models that achieve state-of-the-art performance on the Semantic Textual Similarity task within the MTEB benchmark. GATE leverages Matryoshka Representation Learning and a hybrid loss training approach with Arabic triplet datasets for Natural Language Inference, which are essential for enhancing model performance in tasks that demand fine-grained semantic understanding. GATE outperforms larger models, including OpenAI, with a 20-25\% performance improvement on STS benchmarks, effectively capturing the unique semantic nuances of Arabic.
\end{abstract}

\section{Introduction}

Text embeddings drive advances in clustering, information retrieval, and semantic similarity~\cite{reimers2019sentence, gao2023retrieval, asai2023retrieval, gao2021simcse}. These models aim to map textual information into dense, low-dimensional vector representations that preserve nuanced semantic and contextual relationships. At the heart of many highly effective embedding models lies contrastive learning, a paradigm that optimizes the quality of representation by pulling semantically similar (positive) samples closer while pushing dissimilar (negative) samples apart~\cite{gao2021simcse, he2020momentum, radford2021learning}.

Despite the versatility and success of contrastive learning, most existing text embedding pipelines rely on a two-stage pre-train-fine-tuning process: weakly supervised large-scale pre-training followed by fine-tuning on high-quality text pairs acquired through data mining or manual annotation~\cite{li2023towards, wang2022text, xiao2023c}. Although effective, this approach often relies on the standard InfoNCE loss with in-batch negative samples~\cite{he2020momentum}, achieving robust representations predominantly by using large batch sizes and numerous negative samples. However, InfoNCE alone is not sufficient for all downstream tasks. In particular, sentence-level tasks such as Semantic Textual Similarity (STS) have been shown to benefit less from InfoNCE-based training, indicating a limitation in capturing fine-grained similarity cues~\cite{huang2024piccolo2}. Likewise, key NLP tasks such as STS and classification have yet to be thoroughly integrated into general embedding training objectives.

Arabic presents specific linguistic challenges that complicate Semantic Textual Similarity (STS) tasks. Although Arabic is the fourth most used language on the Internet~\cite{li2018word} and the fifth most spoken language worldwide~\cite{bourahouat2024word}, high-quality Arabic text embeddings are scarce. This scarcity exacerbates issues arising from Arabic's rich morphological structure, characterized by a root-and-pattern system that generates a multitude of derivations, and its flexible syntax, where variable word orders can obscure semantic parallels. Additionally, the frequent omission of diacritics in written Arabic leads to significant ambiguity, as identical word forms may convey different meanings in context. These challenges collectively restrict the accurate capture of semantic nuances, making STS tasks particularly demanding for Arabic NLP applications.

This paper tackles these issues by introducing GATE, a General Arabic Text Embedding model designed to excel in semantic textual similarity and other downstream tasks. Our approach integrates Matryoshka Representation Learning (MRL)~\cite{kusupati2022matryoshka} with a multitask hybrid loss training method. More specifically, we exploit various loss functions tailored to different task objectives—e.g., cosine similarity-based loss for STS and classification-oriented loss for downstream classification tasks. GATE improves semantic distinction by leveraging hard negative datasets and a flexible embedding structure. It addresses the limitations of single-loss approaches like InfoNCE-only training~\cite{gutmann2010noise}.

\vspace{0.3cm}
\textbf{Our contributions} are the following:

\begin{itemize}
  \item \textbf{Hybrid Loss Strategy:} We propose a hybrid loss combining cosine similarity for semantic tasks and softmax-based classification, improving Arabic textual similarity beyond standard InfoNCE.
  \item \textbf{Enhanced Model Robustness:} We incorporate curated Arabic NLI triplet and labeled pair datasets, capturing nuanced semantic relationships crucial for downstream tasks.
  \item \textbf{Scalable Arabic Embeddings:} We adapt Matryoshka Representation Learning to Arabic, enabling efficient multi-dimensional embeddings (768, 512, 256, 128, and 64) with strong performance across tasks.
\end{itemize}

Our models, and data for reproducibility are publicly available~\href{https://huggingface.co/collections/Omartificial-Intelligence-Space/arabic-matryoshka-embedding-models-666f764d3b570f44d7f77d4e}{GATE Collection}.

The paper is organized as follows. Section~\ref{relatedwork} reviews related work. Section~\ref{framework} covers the proposed GATE framework, including our annotated datasets, Matryoshka embeddings training settings, along with a hybrid loss training approach. Section~\ref{results} presents the experimental results, evaluation, benchmarking, and error analysis.

\section{Related Work}\label{relatedwork}

\begin{table*}[h]
\centering
\resizebox{\textwidth}{!}{
\begin{tabular}{c c c c c c}
\hline
\textbf{Work} & \textbf{Embedding Size} & \textbf{Primary Language Focus} & \textbf{Hybrid Loss} & \textbf{Multi-Dimensional} & \textbf{Semantic-Rich Fine-Tuning} \\ 
\hline
OpenAI Text-Embedding v3~\cite{openai2023embeddings} & 1536 / 3072 & Multilingual & \xmarkred & \checkmark & \xmarkred \\  
E5-Mistral-7B-Instruct~\cite{wang2023improving} & 4096 & English-Focused & \xmarkred & \xmarkred & \xmarkred \\  
Udever-Bloom-1B1~\cite{zhang2023language} & 1536 & Multilingual & \xmarkred & \xmarkred & \xmarkred \\  
AraBERT~\cite{antoun2020arabert} & 768 & Arabic-Specific & \xmarkred & \xmarkred & \xmarkred \\  
MARBERT~\cite{abdul2020arbert} & 768 & Arabic-Specific & \xmarkred & \xmarkred & \xmarkred \\  
LaBSE~\cite{feng2020language} & 768 & Multilingual & \xmarkred & \xmarkred & \xmarkred \\  
Multilingual E5~\cite{wang2024multilingual} & 384 & Multilingual & \xmarkred & \xmarkred & \xmarkred \\  
\textbf{GATE Models (Proposed)} & \textbf{768, 512, 256, 128, 64} & \textbf{Semantic Arabic-Specific} & \textbf{\checkmark} & \textbf{\checkmark} & \textbf{\checkmark} \\  
\hline
\end{tabular}
}
\caption{Comparison of GATE with existing models by loss type, embedding dimensions, and training.}
\label{tab:related_works_comp}
\end{table*}

Semantic Textual Similarity (STS)~\cite{cer2017semeval} is a fundamental task in Natural Language Processing (NLP) that measures how closely two sentences align in meaning. Unlike binary classification tasks such as textual entailment or paraphrase detection, STS provides a graded measure of semantic equivalence~\cite{zhao2024enhancing}. It serves as a cornerstone for various NLP applications, including machine translation~\cite{pathak2019english}, text summarization~\cite{liu2022key}, and question answering~\cite{wu2021novel}, making it a crucial benchmark for evaluating embedding models.

Matryoshka Representation Learning (MRL) has emerged as an innovative approach to enhancing text embeddings by introducing hierarchical embedding representations, enabling models to capture multiple fidelity levels while optimizing computational efficiency~\cite{kusupati2022matryoshka}. By dynamically encoding information across varying dimensions, MRL reduces storage requirements and computational overhead without compromising accuracy. Recent advancements, such as OpenAI's text-embedding v3~\cite{openai2024textembeddingv3}, have demonstrated the effectiveness of MRL in semantic representation learning, influencing modern embedding architectures~\cite{koenig2024fluffyembeddings, lee2024gecko, infgrad2024stellamrl}.

Large language models (LLMs) have significantly advanced text embeddings, leveraging massive parameter spaces for complex semantic representations. Models such as \textit{E5-Mistral-7B-Instruct}~\cite{wang2023improving} and \textit{Udever-Bloom-1B1}~\cite{zhang2023language} enhance generalization across domains but remain predominantly optimized for English. OpenAI’s third-generation embedding models~\cite{openai2023embeddings} offer strong multilingual performance but are computationally expensive and lack adaptability for Arabic-specific tasks.

In Arabic NLP, models like \textit{AraBERT} and \textit{MARBERT} have improved language understanding by transitioning from masked language models (MLMs) to sentence embeddings~\cite{reimers-2019-sentence-bert}. While \textit{AraBERT} focuses on formal Arabic~\cite{antoun2020arabert}, \textit{MARBERT} extends coverage to dialectal Arabic through large-scale pretraining~\cite{abdul2020arbert}.

Multilingual models such as \textit{LaBSE}, \textit{SBERT}, and \textit{Multilingual E5} aim to bridge cross-lingual gaps, supporting over 100 languages. However, they struggle with fine-grained Arabic semantics, particularly in STS tasks~\cite{wang2024multilingual}.

To contextualize GATE’s advancements, Table~\ref{tab:related_works_comp} presents a comparative analysis of various text embedding models based on key features, including loss type, embedding dimensionality, fine-tuning methodology, and language specialization.
As shown in Table~\ref{tab:related_works_comp}, most existing models lack hybrid loss strategies, rely on fixed-dimensional embeddings, and are either multilingual or English-centric, making them suboptimal for fine-grained Arabic NLP tasks. GATE addresses these gaps by integrating hybrid loss training, leveraging Matryoshka embeddings, and fine-tuning Arabic semantic datasets, setting a new benchmark for Arabic text embeddings.

\section{GATE Framework}\label{framework}
The GATE framework focuses on Matryoshka representation learning and a multi-task hybrid training approach to enhance Arabic text embeddings. Utilizing the Arabic versions of the Stanford Natural Language Inference (SNLI) and Multi Natural Language Inference (MultiNLI) datasets refines embeddings for optimal performance across various NLP tasks.

\subsection{Dataset}
Our study utilizes Arabic-adapted subsets derived from the Stanford Natural Language Inference (SNLI)~\cite{bowman2015large} and MultiNLI datasets~\cite{kim2019semantic}, originally designed for natural language inference (NLI)~\cite{maccartney2008modeling} tasks. Table~\ref{tab:tab1} summarizes the composition of the Arabic proposed dataset.
 
\begin{table}[ht]
\centering
\renewcommand{\arraystretch}{1.2} 
\small 
\resizebox{1.0\columnwidth}{!}{ 
  \begin{tabular}{lccc} 
    \hline
    \textbf{Subset} & \textbf{Columns} & \textbf{Training} & \textbf{Test} \\
    \hline
    STS & text, text pair, score & 8.63K  & 1.68K \\
    Triplet & text, text triplet & 571K & 6.58K \\
    Pair Classification & text, text pair, label & 981K & 19.7K \\
    \hline
  \end{tabular}
}
\caption{Overview of datasets used in training and test.}
\label{tab:tab1}
\end{table}

As shown in Table~\ref{tab:tab1}, the primary datasets used in this study include the Triplet Subset (571K training, 6.58K test) for contrastive learning, the STS Subset (8.63K training, 1.68K test) for semantic textual similarity evaluation, and the Pair Classification Subset (981K training, 19.7K test) for entailment, neutral, and contradiction classification in hybrid loss training. To adapt NLI datasets for Arabic, we used Neural Machine Translation (NMT)~\cite{klein2017opennmt} with CTranslate2, applying SentencePiece tokenization for efficient processing. Manual reviews ensured high translation accuracy.

\subsection{Proposed Arabic Matryoshka Models}

We introduce a diverse set of Matryoshka-based models optimized for Arabic semantic similarity and natural language inference (NLI). These models enhance representation learning by leveraging hybrid loss training and Matryoshka loss, refining embeddings across different Arabic linguistic contexts.

At the core of our framework is \textit{GATE-AraBERT-V1}, a multi-task trained Arabic embedding model fine-tuned on AllNLI and STS datasets. It is derived from \textit{Arabic-Triplet-Matryoshka-V2}, which extends \textit{AraBERT} using Matryoshka loss and triplet-based training, significantly improving Arabic sentence representations.

Other key models include \textit{Arabic-all-nli-triplet-Matryoshka}, derived from \textit{paraphrase-multilingual-mpnet-base-v2}, optimized for Arabic NLI through triplet learning. \textit{Arabic-labse-Matryoshka} enhances LaBSE’s cross-lingual embeddings for Arabic, while \textit{MARBERT-all-nli-triplet-Matryoshka} adapts MARBERT for both MSA and dialectal Arabic. Finally, \textit{E5-all-nli-triplet-Matryoshka}, built upon multilingual-E5-small, serves as a comparative benchmark for triplet-based learning in Arabic.

Matryoshka models provide a cost-effective alternative to large-scale models like OpenAI’s embeddings, which face scalability and computational challenges. While larger models excel in multilingual tasks, they struggle with fine-grained Arabic semantics. By adapting Arabic and multilingual base models within the Matryoshka framework and leveraging triplet-based training, these models achieve enhanced semantic understanding, improving similarity and NLI tasks while maintaining a balance between cross-lingual adaptability and Arabic linguistic precision~\cite{nacar2024enhancing}.

\subsubsection{Matryoshka Embedding Training Approach}

\begin{figure*}[ht]
 \centering
  \includegraphics[width=1.0\textwidth]{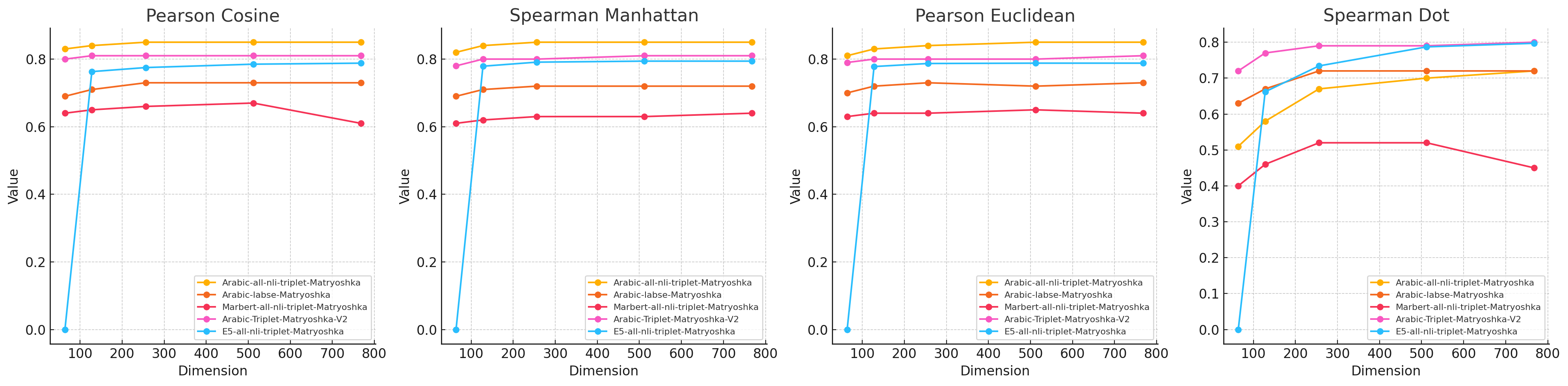}
  \caption{Results of Correlation-based Similarity Metrics on our proposed models.}
  \label{fig:dimensionality}
\end{figure*}

Matryoshka Embedding Models~\cite{kusupati2022matryoshka} introduce an advanced technique for generating adaptable and multi-granular embeddings in natural language processing tasks. These models are designed to capture varying levels of granularity within the embedding vectors, which allows for nuanced representation and efficient computational resource management. This is particularly beneficial in large-scale and resource-constrained scenarios, such as Arabic NLP.

MRL process involves generating a high-dimensional vector $z \in R^{d}$ for each data point $x$ using a deep neural network $F\left(.;\theta_{F} \right)$ parameterized by learnable weights $\theta_{F}$. The key objective of MRL is to ensure that each subset of the first $m$ dimensions of this vector, denoted $z_{1:m} \in R^{m}$ can independently represent the data point effectively. The granularity of the embeddings is controlled through a set of dimensions $M$, which are selected by progressively halving the vector size until reaching a minimal informative state. This approach guarantees that representations remain useful even when truncated to smaller dimensions.

Given a labeled dataset $D = \left\{ \left( x_{1} , y_{1} \right) , ...,\left( x_{N}, y_{N} \right) \right\}$ where $x_{i} \in \chi$ is an input point and $y_{i} \in \left[ L \right]$ is its label, MRL optimizes the multi-class classification loss for each dimension subset $m\in M$. The overall optimization objective is expressed in equation~\ref{eq:Loss_MEL}:

\begin{equation} \label{eq:Loss_MEL}
\mathcal{L}_{\text{MRL}} = \sum_{m \in M} c_m \mathcal{L}_{\text{CE}}(\mathbf{W}^{(m)} \mathbf{z}_{1:m}, y)
\end{equation}

\noindent where \( \mathcal{L}_{\text{MRL}} \) is the MRL loss. \( c_m \) represents the relative importance of each dimension \( m \). \( \mathcal{L}_{\text{CE}} \) denotes the multi-class softmax cross-entropy loss function. \( \mathbf{W}^{(m)} \in \mathbb{R}^{L \times m} \) are the weights of the linear classifier for dimension \( m \). \( \mathbf{z}_{1:m} \in \mathbb{R}^m \) is the truncated embedding vector up to dimension \( m \). \( y \) is the true label corresponding to the input \( x \).

To optimize memory usage, we implement weight-tying across all linear classifiers, setting $\textbf{W}^{\left( m \right)} = \textbf{W}_{1:m}$ for a set of common weights $\textbf{W}$. This variant, known as Efficient MRL, helps manage the memory footprint, which is crucial for handling extensive output spaces.

For the training of Matryoshka models, we utilized the \textit{arabic-nli-triplet} dataset, consisting of 558k triplets, and configured the models to use embeddings at varying dimensions [768, 512, 256, 128, 64]. The training involved using \textit{MultipleNegativesRankingLoss} combined with \textit{MatryoshkaLoss} to handle multiple dimensions effectively. Models were trained on an A100 GPU with a batch size of 128 and a maximum sequence length of 512 tokens. Training configurations and results are managed using the \textit{SentenceTransformerTrainer}.

\subsubsection{Hybrid Loss Training Approach}
A multi-task hybrid loss method has been employed to address limitations in traditional training approaches for embedding models. The training process for our hybrid loss approach was implemented using a multi-dataset strategy that simultaneously leverages both classification and similarity-based objectives. To accommodate the distinct nature of the tasks, we defined two specialized loss functions. For the pair classification task, which involves labeling premise-hypothesis pairs into one of three classes (entailment, neutral, or contradiction), we use a SoftmaxLoss. This loss operates on the sentence embedding dimension extracted from our model and is parameterized by the number of labels (set to 3 in our case). For each premise \( x \), its corresponding hypothesis \( y^+ \) with the correct label (entailment, contradiction, or neutral) is treated as a positive pair, while hypotheses with incorrect labels \( y^- \) are treated as negative pairs. The classification loss function is defined in equation~\ref{eq:cls}:

\begin{equation}\label{eq:cls}
L_{\text{cls}} = -\frac{1}{n} \sum_{i=1}^{n} \log \frac{e^{s(x_i, y^+) / \tau}}{e^{s(x_i, y^+) / \tau} + \sum_{j=1}^{k} e^{s(x_i, y_j^-) / \tau}}
\end{equation}

where \( s(x, y) \) denotes the similarity between the premise \( x \) and the hypothesis \( y \), and \( \tau \) is the temperature scaling parameter. In this case, label-based negatives are applied rather than in-batch negatives. 

For the STS task, which requires capturing subtle semantic differences between sentence pairs, we adopt a cosine similarity-based loss (CoSENTLoss) that effectively penalizes deviations in the computed cosine similarity. The losses are mapped to their respective datasets in a dictionary, ensuring that the appropriate loss function is applied during each training iteration. The cosine similarity loss function is shown in equation~\ref{eq:Lsts}:

\begin{equation}\label{eq:Lsts}
\resizebox{0.95\columnwidth}{!}{$
L_{\text{sts}} = \log \left( 1 + \sum_{s(x_{i}, x_{j}) > s(x_{m}, x_{n})} \exp \left( \frac{\cos(x_m, x_n) - \cos(x_i, x_j)}{\tau} \right) \right)
$}
\end{equation}

where \( \tau \) is the temperature scaling parameter, and \( \cos(\cdot) \) represents the cosine similarity function. \( x_m, x_n, x_i, x_j \) are embeddings of the text pairs.

Training is carried out using a \textit{SentenceTransformerTrainer} configured with meticulously tuned hyperparameters to ensure robust and efficient convergence. In our setup, the training is executed for five epochs with a per-device batch size of 64 and a learning rate of 2e-5, complemented by a warmup ratio of 0.1 to gradually ramp up the learning rate at the onset of training. Frequent logging, evaluation, and checkpointing—executed every 200 steps—enable real-time monitoring and allow for prompt adjustments during training. The final multi-task loss function is formulated in equation~\ref{eq:loss}:

\begin{equation}\label{eq:loss}
L = \begin{cases} 
L_{\text{cls}} & \text{if the task is classification}, \\
L_{\text{sts}} & \text{if the task is STS}. \\
\end{cases}
\end{equation}

This hybrid loss approach ensures that our embedding models are optimally tuned for both classification and STS tasks, thereby enhancing their capability to capture the intricate semantic nuances of Arabic. 

\begin{table*}[!ht]
\centering
\begin{tabular}{ccccccc}
\hline
\textbf{Model}                          & \textbf{Dim} & \textbf{\# Params.} & \textbf{STS17} & \textbf{STS22} & \textbf{STS22-v2} & \textbf{Average} \\
\hline
\textbf{Arabic-Triplet-Matryoshka-V2}       & 768          & 135M               & \textbf{85.31} & \textbf{60.7}  & \textbf{63.96}    & \textbf{69.99}   \\
\textbf{GATE-AraBert-v1}       & 768          & 135M               & 82.78	&59.75	&63.09	& \textbf{68.54}   \\
bert-base-arabertv02                     & 768          & 135M               & 54.53        & 46.86        & 49.95         & 50.45   \\  
\hline
\textbf{Marbert-all-nli-triplet-Matryoshka}       & 768          & 163M               & \textbf{82.18} & \textbf{58.08} & \textbf{61.32}    & \textbf{67.19}   \\
MARBERTv2                                & 768          & 163M               & 60.98        & 49.92        & 53.75         & 54.88   \\  
\hline
\textbf{Arabic-labse-Matryoshka}                  & 768          & 471M               & \textbf{82.46} & 57.25 & 60.58   & \textbf{66.76}   \\
LaBSE                                    & 768          & 471M               & 69.07        & \textbf{57.66}        & \textbf{60.98}        & 62.57   \\  
\hline
\textbf{E5-all-nli-triplet-Matryoshka}            & 384          & 278M               & \textbf{80.37} & 56.34 & 59.64   & \textbf{65.45}   \\
multilingual-e5-small                    & 384          & 278M               & 74.62        & \textbf{58.13}       & \textbf{61.4}          & 64.72   \\  
\hline
\textbf{Arabic-all-nli-triplet-Matryoshka}        & 768          & 135M               & \textbf{82.4}  & 51.38 & 54.45    & \textbf{62.74}   \\
paraphrase-multilingual-mpnet-base-v2    & 768          & 135M               & 79.09        & \textbf{52.18}        & \textbf{55.37}        & 62.21   \\  
\hline
\end{tabular}
\caption{Performance comparison of Matryoshka models vs. their base counterparts on MTEB benchmarks.}
\label{tab:tab2}
\end{table*}

\section{Results and Discussion}\label{results}

\subsection{Results of Correlation Similarity Metrics}

In order to assess the robustness of Matryoshka embeddings across different dimensions, we evaluated our Matryoshka models across multiple embedding sizes (768, 512, 256, 128, and 64). We employ correlation-based similarity metrics, commonly used in text embedding evaluations, to measure the consistency of embeddings across different dimensions. Figure~\ref{fig:dimensionality} presents the results using Pearson and Spearman correlation metrics, computed with different distance functions: Cosine, Manhattan, Euclidean, and Dot Product.

As shown in Figure~\ref{fig:dimensionality}, higher-dimensional embeddings (768, 512) consistently achieve superior performance, while lower-dimensional embeddings (128, 64) exhibit a noticeable decline, particularly in dot product-based similarity measures. \textit{Arabic-all-nli-triplet-Matryoshka} achieves the highest scores across Pearson Cosine, Spearman Manhattan, and Pearson Euclidean, maintaining values around 0.85 for larger dimensions. \textit{Arabic-Triplet-Matryoshka-V2} follows closely with stable performance across all metrics, scoring approximately 0.80 at higher dimensions. \textit{Arabic-labse-Matryoshka} remains robust, averaging 0.72–0.73, while \textit{Marbert-all-nli-triplet-Matryoshka} shows slightly lower results, particularly in Spearman Dot and Pearson Cosine (0.61–0.67). \textit{E5-all-nli-triplet-Matryoshka} demonstrates a declining trend, especially in Spearman Dot at lower dimensions. These findings reinforce the trade-off between STS accuracy and embedding efficiency, emphasizing the importance of selecting an optimal embedding size based on computational constraints and task requirements.

\subsection{Performance Evaluation on Arabic STS MTEB Benchmarks}

To evaluate the effectiveness of Matryoshka and Multi-Task Hybrid Loss methods, we conduct experiments on GATE models, and their base counterparts using the Massive Text Embedding Benchmark (MTEB)~\cite{muennighoff2022mteb} for Arabic. MTEB provides a large-scale evaluation across various NLP tasks, including Semantic Textual Similarity (STS), with key Arabic metrics: STS17, STS22, and STS22-v2~\cite{cer2017semeval}. These metrics assess STS on a scale from 0 to 5, focusing on Arabic-Arabic sentence pairs. Table~\ref{tab:tab2} presents the comparative performance of Matryoshka embeddings against their base models.

As shown in Table~\ref{tab:tab2}, Matryoshka-based models consistently outperform their base counterparts. \textit{Arabic-Triplet-Matryoshka-V2} achieves the highest performance (69.99 avg.), excelling in STS17 (85.31), while \textit{GATE-AraBERT-V1} follows closely with 68.54. Interestingly, \textit{GATE-AraBERT-V1}—which incorporates multi-task hybrid loss training—scores slightly lower than \textit{Arabic-Triplet-Matryoshka-V2}, likely due to trade-offs in optimizing multiple objectives (STS and classification). While hybrid loss improves generalizability, Matryoshka loss preserves fine-grained sentence embedding alignment better, explaining this marginal gap.

Among other Matryoshka adaptations, \textit{Marbert-all-nli-triplet-Matryoshka} scores 67.19, showcasing robust performance across STS22 and STS22-v2, while \textit{Arabic-labse-Matryoshka} follows closely with 66.76. The \textit{E5-all-nli-triplet-Matryoshka}, despite using a smaller 384-dimensional embedding space, maintains competitive results with 65.45, demonstrating an effective balance between efficiency and performance.

In contrast, base models significantly underperform, with \textit{bert-base-arabertv02} achieving the lowest score at 50.45 and \textit{paraphrase-multilingual-mpnet-base-v2} reaching 62.21. These findings underscore the effectiveness of Matryoshka Representation Learning (MRL) and hybrid loss strategies in refining Arabic embedding models, enhancing STS understanding, and optimizing performance across Arabic NLP benchmarks.

\begin{table}[!ht]
\centering
\renewcommand{\arraystretch}{1.2} 
\resizebox{\columnwidth}{!}{%
\begin{tabular}{ccccc}
\hline
\textbf{Loss} & \textbf{STS17} & \textbf{STS22} & \textbf{STS22-v2} & \textbf{Average} \\
\hline
$L_{CE}$ & 54.53 & 46.86 & 49.95 & 50.45 \\
$L_{MRL}$ & \textbf{85.31} & \textbf{60.7} & \textbf{63.96} & \textbf{69.99} \\
$L_{sts} + L_{cls}$ & 82.78 & 59.75 & 63.09 & 68.54 \\
\hline
\end{tabular}%
}
\caption{Effect of Matryoshka and hybrid loss functions on Arabic STS benchmarks}
\label{tab:tab3}
\end{table}

\begin{figure*}[ht]
 \centering
  \includegraphics[width=1.0\textwidth]{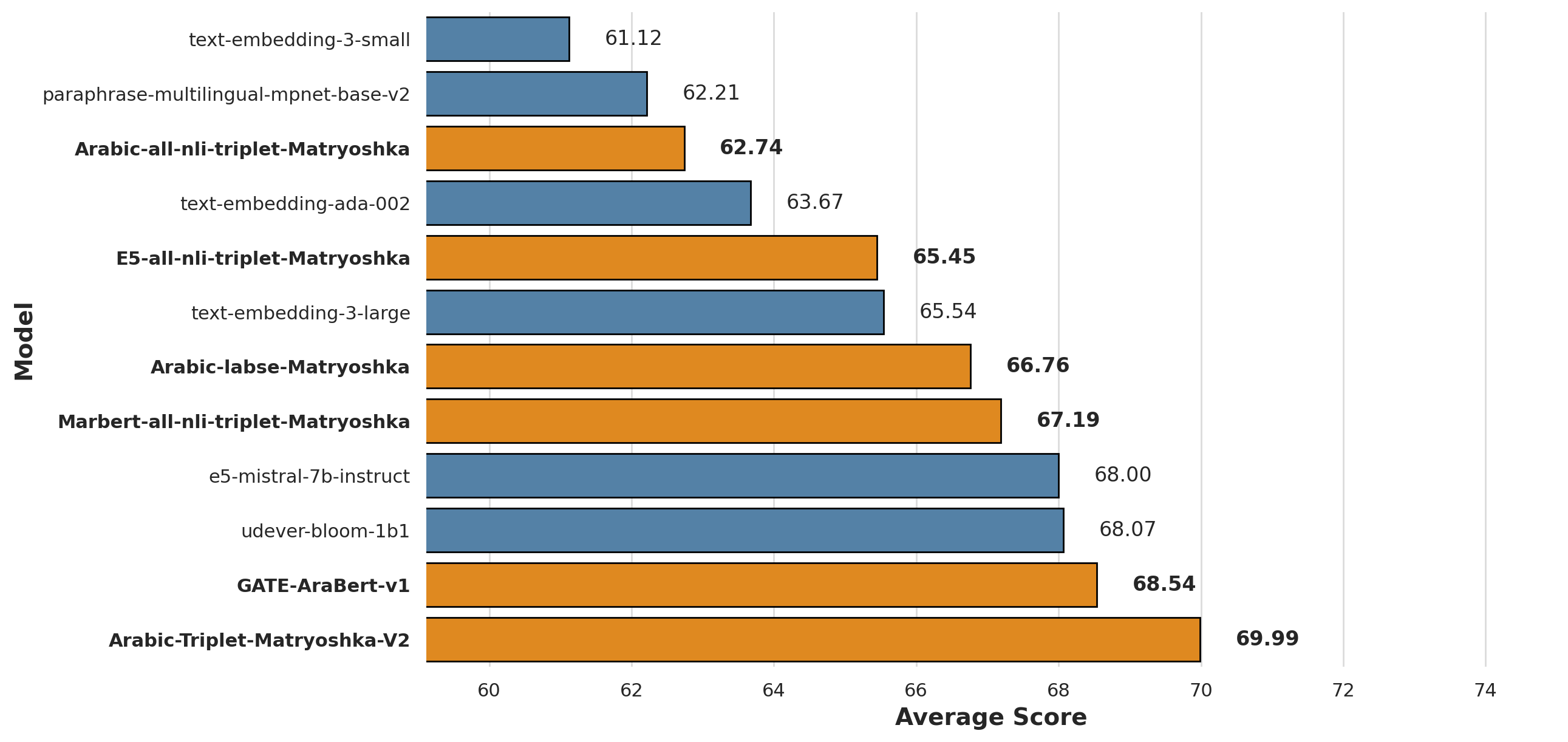}
  \caption{Performance comparison between Matryoshka models and larger models on MTEB Arabic benchmarks.}
  \label{fig:1}
\end{figure*}

Table~\ref{tab:tab3} highlights the impact of different loss functions on the best-performing models, \textit{Arabic-Triplet-Matryoshka-V2} and \textit{GATE-AraBERT-V1}, across the three Arabic STS benchmarks in MTEB. The results demonstrate the crucial role of loss selection in optimizing model performance for STS tasks.

As shown in Table~\ref{tab:tab3}, the baseline cross-entropy loss $L_{CE}$ yields the lowest average score of 50.45, reinforcing its limitations in learning high-quality embeddings for fine-grained STS. In contrast, \textit{Arabic-Triplet-Matryoshka-V2}, trained with Matryoshka loss $L_{MRL}$, achieves the highest performance with an average of 69.99, significantly improving on STS17 equal to 85.31. Similarly, the hybrid loss approach ($L_{sts} + L_{cls}$), applied to \textit{GATE-AraBERT-V1}, achieves a strong performance with an average of 68.54. While slightly lower than MRL, this result highlights the trade-off between generalization and fine-tuned similarity alignment. Hybrid loss optimizes embeddings for both STS and classification tasks, making it more versatile across different NLP applications. 

Moreover, the effectiveness of MRL extends beyond performance gains. It enables models to retain their high-level semantic understanding even when embeddings are trained at progressively smaller dimensions, reducing computational and memory costs without significant degradation in performance. This characteristic is particularly beneficial in resource-constrained settings, where maintaining efficiency without sacrificing accuracy is critical. Table~\ref{tab:tab4} shows the performance of the best-performing model, \textit{Arabic-Triplet-Matryoshka-V2}, across various embedding dimensions (768, 512, 256, 128, and 64) on STS MTEB metrics.

\begin{table*}[!ht]
\centering
\renewcommand{\arraystretch}{1.2} 
\begin{tabular}{ccccc}
\hline
\textbf{Evaluation Dim.} & \textbf{STS17} & \textbf{STS22} & \textbf{STS22-v2} & \textbf{Average} \\
\hline
768       & 85.31 & \textbf{60.7}  & 63.96    & \textbf{69.99}   \\
512       & 85.17 & 60.62 & \textbf{63.98}    & 69.92   \\
256       & \textbf{85.39} & 60.41 & 63.77    & 69.86   \\
128       & 84.67 & 60.27 & 63.62    & 69.52   \\
64        & 84.04 & 60.44 & 63.8     & 69.43   \\
\hline
\end{tabular}%
\caption{Impact of embedding dimensions on the performance of \textit{Arabic-Triplet-Matryoshka-V2}.}
\label{tab:tab4}
\end{table*}

As shown in Table~\ref{tab:tab4}, results demonstrate that the model maintains robust performance across all dimensions. At the full 768-dimensional embedding, the model achieves an average score of 69.99, with 85.31 on STS17. Even when reduced to 512 and 256 dimensions, the performance remains nearly unchanged, with average scores of 69.92 and 69.86, respectively. Even at the lowest dimension of 64, the model still delivers a strong average score of 69.43, confirming that MRL allows for significant compression without substantial loss in accuracy. 

\subsection{Comparison of GATE Models with LLMs}

To assess the efficiency of GATE models, we conducted a comparative evaluation against larger models, including \textit{e5-mistral-7b-instruct} (7B parameters), \textit{udever-bloom-1b1} (1B parameters), and OpenAI’s \textit{text-embedding-3-small/large} and \textit{text-embedding-ada-002}. Figure~\ref{fig:1} highlights how Matryoshka models, despite smaller sizes, outperform or match billion-parameter LLMs in Arabic STS tasks.

As shown in Figure~\ref{fig:1}, the \textit{Arabic-Triplet-Matryoshka-V2} model and \textit{GATE-Arabert-V1}, with only 135M parameters, achieved the highest scores of 69.99 and 68.54 respectively, surpassing both \textit{e5-mistral-7b-instruct} (68.00) and \textit{udever-bloom-1b1} (68.07), despite their significantly larger parameter sizes. Similarly, OpenAI's \textit{text-embedding-ada-002} achieved a lower average score of 63.67, while the larger \textit{text-embedding-3-large} model reached 65.54. Other Matryoshka models, such as \textit{Marbert-all-nli-triplet-Matryoshka} and \textit{Arabic-labse-Matryoshka}, demonstrated competitive performance, achieving 67.19 and 66.76, respectively. These results underscore the efficiency of the Matryoshka framework, demonstrating that smaller, well-optimized models can achieve state-of-the-art performance in STS tasks without the need for billions of parameters.

\subsection{Error Analysis}

We conducted an error analysis on Arabic-trained Matryoshka models by comparing their predictions against ground truth labels across high, moderate, and low similarity categories. This evaluation highlights patterns of overestimation and underestimation, particularly in distinguishing semantically unrelated pairs, as shown in Tables~\ref{tab:tab5}, \ref{tab:tab6}, and \ref{tab:tab7}.

As observed in the no similarity case in Table~\ref{tab:tab6}, most models assigned considerably higher similarity scores than the ground truth of 0.1, with some exceeding 0.4, indicating a false positive bias. This suggests that while models effectively recognize shared words, they may struggle to distinguish true semantic relationships when there is lexical overlap. Notably, \textit{GATE-AraBERT-V1} achieved the most accurate prediction with a score of 0.04, indicating that its hybrid loss training aids in learning better distinctions between semantically unrelated sentences.

For moderate similarity pairs in Table~\ref{tab:tab5}, models exhibit better alignment with ground truth, with scores ranging between 0.66 and 0.83, reinforcing their robustness in handling nuanced semantic relationships. \textit{GATE-AraBERT-V1} slightly overestimates the similarity with a score of 0.81, while \textit{Marbert-all-nli-triplet-Matryoshka} and \textit{Arabic-labse-Matryoshka} reach the highest scores at 0.836 and 0.835, respectively.

For high similarity cases shown in Table~\ref{tab:tab7}, all models perform well, scoring above 0.84, closely mirroring the ground truth of 1.0. However, \textit{GATE-AraBERT-V1} achieves a slightly lower score of 0.73, suggesting that hybrid loss training may introduce more conservative similarity estimations compared to Matryoshka loss models.

\begin{table}[ht]
\centering
\renewcommand{\arraystretch}{1.5} 
\resizebox{\columnwidth}{!}{%
\begin{tabular}{|l|l|p{0.35\textwidth}|p{0.35\textwidth}|}
\hline
\textbf{\Large Model} & \textbf{\Large Score} & \textbf{\Large Sentence1} & \textbf{\Large Sentence2} \\ \hline
\multirow{2}{*}{\Large Ground Truth} & \multirow{2}{*}{\Large 0.1} & \multirow{7}{*}{\Large \begin{tabular}[c]{@{}c@{}} \RL{رجل يعزف على الجيتار} \\ (A man playing the guitar) \end{tabular}} & \multirow{7}{*}{\Large \begin{tabular}[c]{@{}c@{}} \RL{رجل يقود سيارة} \\ (A man driving a car) \end{tabular}} \\
 &  &  &  \\ \cline{1-2}
\Large Arabic-all-nli-triplet-Matryoshka & \Large 0.48 &  &  \\ \cline{1-2}
\Large Arabic-Triplet-Matryoshka-V2 & \Large 0.48 &  &  \\ \cline{1-2}
\Large GATE-AraBert-v1 & \Large 0.04 &  &  \\ \cline{1-2}
\Large Arabic-labse-Matryoshka & \Large 0.32 &  &  \\ \cline{1-2}
\Large Marbert-all-nli-triplet-Matryoshka & \Large 0.38 &  &  \\ \hline
\end{tabular}%
}
\caption{Model scores for a no similarity sample.}
\label{tab:tab6}
\end{table}

\begin{table}[ht]
\centering
\renewcommand{\arraystretch}{1.5} 
\resizebox{\columnwidth}{!}{%
\begin{tabular}{|l|l|p{0.35\textwidth}|p{0.35\textwidth}|}
\hline
\textbf{\Large Model} & \textbf{\Large Score} & \textbf{\Large Sentence1} & \textbf{\Large Sentence2} \\ \hline
\multirow{2}{*}{\Large Ground Truth} & \multirow{2}{*}{\Large 0.72} & \multirow{7}{*}{\Large \begin{tabular}[c]{@{}c@{}} \RL{الرجال يلعبون كرة القدم} \\ (men are playing football) \end{tabular}} & \multirow{7}{*}{\Large \begin{tabular}[c]{@{}c@{}} \RL{الأولاد يلعبون كرة القدم} \\ (boys are playing football) \end{tabular}} \\
 &  &  &  \\ \cline{1-2}
\Large Arabic-all-nli-triplet-Matryoshka & \Large 0.685 &  &  \\ \cline{1-2}
\Large Arabic-Triplet-Matryoshka-V2 & \Large 0.661 &  &  \\ \cline{1-2}
\Large GATE-AraBert-v1 & \Large 0.81 &  &  \\ \cline{1-2}
\Large Arabic-labse-Matryoshka & \Large 0.835 &  &  \\ \cline{1-2}
\Large Marbert-all-nli-triplet-Matryoshka & \Large 0.836 &  &  \\ \hline
\end{tabular}%
}
\caption{Model scores for a moderate similarity sample.}
\label{tab:tab5}
\end{table}

\begin{table}[ht]
\centering
\renewcommand{\arraystretch}{1.5} 
\resizebox{\columnwidth}{!}{%
\begin{tabular}{|l|l|p{0.35\textwidth}|p{0.35\textwidth}|}
\hline
\textbf{\Large Model} & \textbf{\Large Score} & \textbf{\Large Sentence1} & \textbf{\Large Sentence2} \\ \hline
\multirow{2}{*}{\Large Ground Truth} & \multirow{2}{*}{\Large 1} & \multirow{7}{*}{\Large \begin{tabular}[c]{@{}c@{}} \RL{رجل يقوم بخدعة بالبطاقات} \\ (A man doing a \\ card trick) \end{tabular}} & \multirow{7}{*}{\Large \begin{tabular}[c]{@{}c@{}} \RL{رجل يقوم بخدعة ورق} \\ (A man performing a \\card trick) \end{tabular}} \\
 &  &  &  \\ \cline{1-2}
\Large Arabic-all-nli-triplet-Matryoshka & \Large 0.91 &  &  \\ \cline{1-2}
\Large Arabic-Triplet-Matryoshka-V2 & \Large 0.87 &  &  \\ \cline{1-2}
\Large Arabic-labse-Matryoshka & \Large 0.84 &  &  \\ \cline{1-2}
\Large Marbert-all-nli-triplet-Matryoshka & \Large 0.85 &  \\ \cline{1-2}
\Large GATE-AraBert-v1 & \Large 0.73 &  &  \\ \hline
\end{tabular}%
}
\caption{Model scores for a high similarity sample.}
\label{tab:tab7}
\end{table}

\section{Limitations}

This work presents certain limitations. The lack of comprehensive Arabic NLP benchmarks restricts a broader evaluation beyond STS tasks. Additionally, error analysis reveals a tendency to overestimate similarity in unrelated sentence pairs, often due to shared lexical elements, leading to false positives. Enhancing negative pair handling could further refine model accuracy. While our approach is optimized for Arabic, the methodology holds the potential for multilingual adaptation, expanding its applicability. 

\section{Conclusion}\label{conclusion}

In this work, we introduced GATE, a General Arabic Text Embedding model leveraging MRL and hybrid loss training to enhance STS tasks. Evaluations on MTEB benchmarks confirmed strong performance retention across reduced dimensions and improved generalization over larger models. GATE fills key gaps in Arabic NLP by optimizing embeddings for fine-grained Arabic semantics. Future work will extend Arabic NLP benchmarks, diversify datasets, and explore multilingual generalization for broader real-world impact.

\section*{Acknowledgments}

The authors thank Prince Sultan University for their support.

\bibliography{custom}

\end{document}